\documentclass[10pt,twocolumn,letterpaper]{article}

\usepackage{iccv}
\usepackage{times}
\usepackage{epsfig}
\usepackage{graphicx}
\usepackage{amsmath}
\usepackage{amssymb}

\usepackage{subfiles}
\usepackage{subfig}
\usepackage{boldline} 
\usepackage{dsfont}
\usepackage{color,soul}
\usepackage{balance}
\usepackage{multirow}
\usepackage{multicol}

\usepackage{setspace}

\usepackage{comment}
\usepackage{bbm}
\newcommand{\mainenc}{\textit{Main Encoder}}
\newcommand{\partenc}{\textit{Partner Encoder}}
\newcommand{\conloss}{\mathcal{L}_{CT}}
\newcommand{\supercon}{SupCT}
\newcommand{\superconloss}{\mathcal{L}_{SupCT}}
\newcommand{\partfunc}{$f_P$}
\newcommand{\mainfunc}{$f_M$}

\usepackage{enumitem}
\usepackage{gensymb}
\setitemize{noitemsep,topsep=0pt,parsep=0pt,partopsep=0pt}
\usepackage[dvipsnames]{xcolor}

\definecolor{fig3_class4}{RGB}{147,116,188}
\definecolor{fig3_class2}{RGB}{99,167,82}
\definecolor{fig3_class1}{RGB}{240,133,54}

\usepackage[pagebackref=true,breaklinks=true,letterpaper=true,colorlinks,bookmarks=false]{hyperref}
\usepackage[capitalise]{cleveref}

\usepackage{AbdAlmageed_style}

\usepackage[breaklinks=true,bookmarks=false]{hyperref}

\iccvfinalcopy %

\def\iccvPaperID{****} %

\ificcvfinal\fi

\begin{document}

\title{Partner-Assisted Learning for Few-Shot Image Classification}

\author{
Jiawei Ma$^{\dagger,1}$ \qquad Hanchen Xie$^{\dagger,2}$ \qquad Guangxing Han$^1$ \qquad Shih-Fu Chang$^1$ \\ Aram Galstyan$^2$ \qquad Wael Abd-Almageed$^2$\\
$^1$ Columbia University \qquad \qquad \qquad
$^2$ USC Information Sciences Institute \\
\tt\small \{jiawei.m, gh2561, sc250\}@columbia.edu, \{hanchenx, galstyan, wamageed\}@isi.edu}

\maketitle
\ificcvfinal\thispagestyle{empty}\fi

\let\thefootnote\relax\footnotetext{$^\dagger$Equal contribution.}
\begin{abstract}
Few-shot Learning has been studied to mimic human visual capabilities and learn effective models without the need of exhaustive human annotation. Even though the idea of meta-learning for adaptation has dominated the few-shot learning methods, how to train a feature extractor is still a challenge.
In this paper, we focus on the design of training strategy to obtain an elemental representation such that the prototype of each novel class can be estimated from a few labeled samples. We propose a two-stage training scheme, Partner-Assisted Learning (PAL), which first trains a \partenc{} to model pair-wise similarities and extract features serving as soft-anchors, and then trains a \mainenc{} by aligning its outputs with soft-anchors while attempting to maximize classification performance. Two alignment constraints from logit-level and feature-level are designed individually. For each few-shot task, we perform prototype classification. Our method consistently outperforms the state-of-the-art methods on four benchmarks. Detailed ablation studies of PAL are provided to justify the selection of each component involved in training. 
\end{abstract}

\section{Introduction}
Deep learning has achieved impressive success in many vision tasks, such as image classification~\cite{alex-net, ILSVRC15, resnet}, object detection~\cite{ren2015faster, redmon2016you, faster-rcnn}, and image segmentation~\cite{FCN, deeplab, Mask-R-CNN}, especially when sufficient labeled data is available for training. However, data annotation can be expensive and large scale annotated data is not always available~\cite{DeepLearning, one-shot, sung2018learning, muscle}.

Few-shot learning has been proposed to mimic human vision systems, which is capable of  learning the visual appearance of new objects  with only a few (\eg, 1 or 5)  instances \cite{one-shot, miniImagenet}. To facilitate few-shot learning for fast model adaptation, meta-learning has been employed to simulate few-shot tasks during training, by either designing an optimal algorithm for adaptation ~\cite{finn2017model,nichol2018reptile} or learning a shared feature space for prototype-based classification~\cite{snell2017prototypical,FC100,li2020boosting}.
\begin{figure}[t]
\begin{center}
   \includegraphics[width=0.95\linewidth]{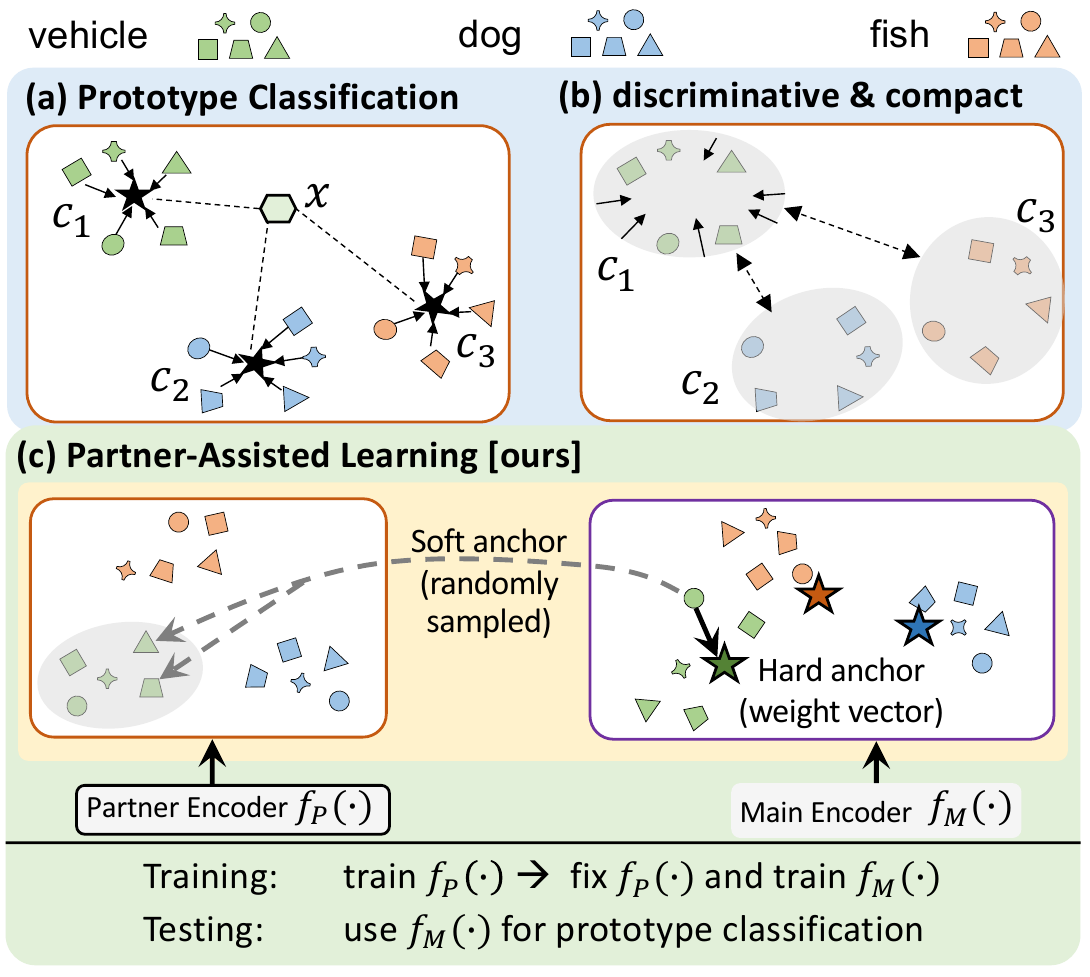}
\end{center}
   \caption{(a) Prototype classification calculates the few-shot prototypes and classifies a sample by comparing its similarity to each prototype. 
   (b) The discriminative feature distribution with compact clusters benefits the prototype classification~\cite{LGM2018, largeMarginSoftmax}.
   (c) We propose a Partner-Assisted Learning framework, in which a pre-trained partner encoder, \partfunc{}, is used to generate soft-anchors to regularize the learning of  main encoder, \mainfunc{}, which will be used at the inference time.}
\label{fig:pal_concept}
\end{figure}
As shown in \cref{fig:pal_concept}(a), prototype classification methods~\cite{chen2020new, dhillon2019baseline, snell2017prototypical, tian2020rethinking} estimate the few-shot prototypes by averaging the features of a few labeled samples (\ie, support).
A new sample (\ie, query) is classified by comparing its cosine similarity with all prototypes using nearest neighborhood search.
As illustrated in \cref{fig:pal_concept}(b), in a classification context, the feature distribution is supposed to be (1) compact within each cluster (\ie, supporting high intra-class similarities), and (2) discriminative between clusters (i.e. supporting large inter-class distances).

Recent work has shown that pretraining a model on a large scale (base) dataset  with full supervision can serve as a strong baseline~\cite{chen2020new} for the novel few-shot tasks by performing prototype classification~\cite{chen2020new, dhillon2019baseline, snell2017prototypical, tian2020rethinking}.
For each base class, conventional fully-supervised pretraining using class labels~\cite{tian2020rethinking,chen2020new}  learns a unique weight vector, which serves as a hard-anchor.
By minimizing the cross-entropy loss with respect to (w.r.t.) the class label, each image feature is pushed towards its corresponding class anchor. Therefore, for each class, the average of features is expected to represent the class during few-shot classification.

The feature extractor pretrained on the base classes for classification may suppress details irrelevant to the base domain~\cite{chen2018zero}, while these details could be  discriminative for novel classes. Thus, 
incorporating instance comparison to preserve details can facilitate few-shot learning on novel domains. 
Knowledge distillation formulates a teacher-student setting and compares the outputs from two models for the same image~\cite{tian2020rethinking}. 
For each image, the teacher model generates soft-labels to model the proximity between different classes. By comparing the outputs of teacher model and student model, the student model is trained with the soft-label so that more details indicating class relationship could be preserved. Thus, the student model achieves higher accuracy on few-shot tasks. 
Despite the success of knowledge distillation, the performance improvement is still limited since the teacher model has once been rigidly optimized according to hard-anchors of base classes.

Besides comparing the outputs of the same instance from two networks using cosine similarity, a single network can be trained for pair-wise comparison, so that its outputs of a few randomly selected support samples can dynamically represent the class center~\cite{snell2017prototypical}.
Metric-based meta-learning methods, such as prototypical network ~\cite{snell2017prototypical}, have been proposed to learn to represent a class by aggregating support features. This way, representative centers are dynamically estimated according to a few labeled data. 
Similarly, supervised contrastive learning~\cite{khosla2020supervised} performs pair-wise comparison, where each feature is sampled from the training set and individually represents the class without aggregation. 

Inspired by the dynamic and individual representative employed in prototypical learning and supervised contrastive learning, to improve the generalization ability of feature extractor, we propose to extract features that can be used to dynamically represent classes, and set those features as soft-anchors to regularize the feature extractor trained with hard-anchors.
Comparing with knowledge distillation, instead of aiming to iterate the feature extractor that has already been optimized w.r.t hard-anchors, our method uses diverse features on the base domain to regularize a new feature extractor which is trained with class label under cross-entropy loss from scratch. The contributions of this paper are as follows:
\begin{itemize}
    \item We propose Partner-Assisted Learning (PAL): a framework for representation learning in few-shot classification setting, in which the \partenc{} and \mainenc{} are trained sequentially such that the features from \partenc{} are used as soft-anchors to regularize the training of \mainenc{} from scratch.
    
    \item We propose two alignment approaches on both feature-level and logit-level, which utilizes the soft-anchors for regularization during training with class labels.

    \item We show that PAL consistently achieves state-of-the-art performance on four few-shot benchmarks, and improves the classification accuracy in a supervised learning setting. We also provide comprehensive ablation studies to justify the design of each component.
\end{itemize}

\section{Related Work}
\noindent \textbf{Prototype Classification} has been widely used in metric-based methods for few-shot classification.
Prototypical networks~\cite{snell2017prototypical} simulate few-shot tasks using episodes during meta-training. In each episode, a few labeled training samples are randomly sampled and then class prototype is estimated by averaging the extracted features. 
The quality of the estimated class prototype is evaluated by classifying the query features. 
Similarly, models trained by the supervised contrastive loss (\supercon)~\cite{khosla2020supervised} learn to maximize the similarity between all instances of the same class so that all instances are clustered together and each class can be represented by every instance feature of that class. 
Furthermore, the concept of meta-learning is also employed to estimate the task-adaptive metric by learning to scale the metric or add margins for prototype classification, which has shown clear benefit on few-shot tasks~\cite{FC100,li2020boosting,QAFewDet2021,han2021meta}.

Recently, networks pretrained with fully-supervised classification tasks \cite{chen2020new} have been treated as strong baselines for few-shot classification. A unique prototype for each class is  learned through class labels, \ie, one-hot vectors, to indicate the discrimination between classes.
Furthermore, RFS~\cite{tian2020rethinking} shows further improvements using knowledge distillation with soft labels based on the conventional network trained with cross-entropy loss.

\noindent \textbf{Regularization to Cross-Entropy (CE) loss}. CE loss~\cite{murphy2012machine} is widely used in  fully-supervised tasks due to its simplicity, where it learns a classification hyperplane in a high-dimensional representation space. Regularization can be added to encourage the intra-class compactness, by setting large margin~\cite{largeMarginSoftmax}. Various loss functions, such as center loss~\cite{center_loss}, L-GM loss~\cite{LGM2018}, and Ring Loss~\cite{ring_loss} have been introduced to emphasize certain embedding distributions in the latent space. In face recognition, triplet loss~\cite{Schroff_2015_CVPR} has widely been used where an image triplet is constructed by sampling a positive pair and a negative pair as anchors. 

\noindent \textbf{Knowledge Transfer between Networks:} 
Knowledge Distillation~\cite{hinton2015distilling} has been proposed to perform uni-direction knowledge transfer, which uses a strong teacher model to train a simple student model on the same task.  Other work on knowledge distillation also show the advantages in semi-supervised learning  ~\cite{meanteacher, simclrv2}. 
The strong teacher model is pretrained with one-hot vectors, and then generates soft-labels for the student model. 
Although the teacher model can serve as a strong baseline, the large negative logits may hurt the distillation process and such logits need to be smoothed by modifying the Softmax operation~\cite{hinton2015distilling}.
Contrarily, mutual learning~\cite{zhang2018deep} studies bi-direction knowledge sharing, where two networks are trained jointly from scratch with the same objective. 
Both uni-direction and bi-direction knowledge transfer have demonstrated  ability to learn a better representation than a single network.

\section{Partner-Assisted Learning}

\begin{figure*}[t]
\begin{center}
   \includegraphics[width=0.95\textwidth]{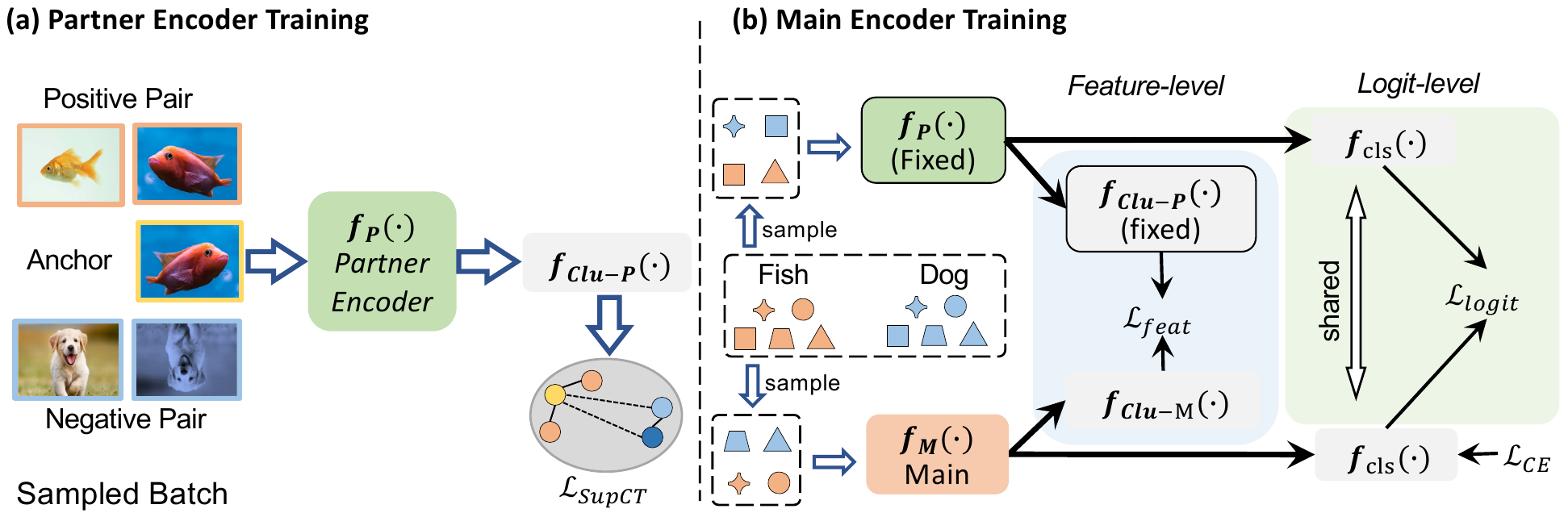}
\end{center}
   \caption{Training pipeline of Partner-Assisted Learning. (a) Train the \partenc{} \partfunc{} by supervised contrastive learning~$\superconloss{}$ to model the pair-wise similarity among all features. (b) Train the \mainenc{} \mainfunc{} by imposing either logit-level or feature-level alignments using the pretrained \partfunc{}. Both \partfunc{} and \mainfunc{} use ResNet-12\cite{resnet} and each gray block denotes a fully-connected layer.
   }
  \label{fig:pal_detail}
\end{figure*}

In this section, we present the proposed Partner-Assisted Learning (PAL) to learn an embedding function. 
As illustrated in \cref{fig:pal_detail}, PAL consists of \textit{Partner} and \textit{Main} Encoders. 
Task formulation and notation are defined in \cref{sec:preliminary}. 
The objective for the \partenc{} is presented in \cref{sec:enc_part}.
The framework of imposing alignment constraints for the \mainenc{} is discussed in \cref{sec:enc_main}.

\subsection{Learning-Task Formulation}\label{sec:preliminary}

In the few-shot learning, we are first given a \textit{base} dataset $\mathcal{D}_{base}$, which consists of abundant amount of labeled samples. All sample labels in $\mathcal{D}_{base}$ belong to the base class set $\mathcal{C}_{base}$. 
Then, we are given a \textit{novel} set $\mathcal{D}_{novel}$, from which each episode $\mathcal{D}_{epi}$ is sampled. All sample labels in $\mathcal{D}_{novel}$ are from the novel class set $\mathcal{C}_{novel}$ where the class sets for \textit{base} and \textit{novel} are disjoint, \ie, $\mathcal{C}_{base} \cap \mathcal{C}_{novel} = \varnothing$. 
Each episode $\mathcal{D}_{epi}=(\mathcal{D}_{\mathcal{S}},\mathcal{D}_{\mathcal{Q}})$ is composed of a support set $\mathcal{D}_{\mathcal{S}}$ for prototype estimation and a query set $\mathcal{D}_{\mathcal{Q}}$ for evaluation. 
For an $N$-way $K$-shot task, the $\mathcal{D}_{\mathcal{S}} \cup \mathcal{D}_{\mathcal{Q}}$ in an episode contains $N$ novel classes drawn from $\mathcal{C}_{novel}$, and $\mathcal{D}_{\mathcal{S}}$ contains $K$ labeled samples for each class.

As illustrated in \cref{fig:pal_detail},  $\mathcal{D}_{base}$ is first used to train the \partenc{} \partfunc{} to generate soft-anchors. Then,  \partfunc{} is fixed and  $\mathcal{D}_{base}$ is used to train the \mainenc{} \mainfunc{}, which is regularized by the alignment constraints on either logit-level or feature-level from \partfunc{} under the PAL framework. 
During the few-shot evaluation, similar to~\cite{tian2020rethinking,chen2020new}, we directly use the pre-trained \mainfunc{} to estimate the prototype of each class using $\mathcal{D}_{\mathcal{S}}$ and classify the testing samples in $\mathcal{D}_{\mathcal{Q}}$.

\subsection{Partner Encoder}\label{sec:enc_part}

A \partenc{} \partfunc{} is trained using supervised contrastive learning (\supercon{}) to do clustering and perform pair-wise comparison among all feature instances. 
The features of the same class are pushed together while the features from different classes are pushed away.
The detail of supervised contrastive learning is presented below. 

\noindent \textbf{Supervised Contrastive Learning:} Given a batch $\mathcal{D}_{\text{raw}}$ with $B$ images, \ie, $|\mathcal{D}_{\text{raw}}| = B$, an augmented batch with $2B$ samples is generated by performing two separate augmentations on each image,
\begin{equation}\label{eq:M_composition}
    \mathcal{D} = \text{Concat}(\text{Aug}(\mathcal{D}_{\text{raw}}),\text{Aug}(\mathcal{D}_{\text{raw}})),
\end{equation}
where $\text{Aug}$ indicates a data augmentation function-group which randomly transforms images.
For each image $\mathcal{D}(i)$ where $i \in \mathcal{I} \equiv \{1...2B\}$, a positive index set $\mathcal{I}_{\text{pos}}(i) \subset \mathcal{I} \setminus \{i\}$ is selected, such that all images $\mathcal{D}(j)$ for $j \in \mathcal{I}_{\text{pos}}(i)$ are of the same class as $\mathcal{D}(i)$. Then, the supervised contrastive loss is defined as 
\begin{align}\label{eq:supCon}
    &\superconloss{}(\mathcal{D}) = \sum_{i\in I}\frac{-1}{|\mathcal{I}_{\text{pos}}(i)|}
    \sum_{j\in \mathcal{I}_{\text{pos}}(i)} \Theta(i,j) \\
    &\Theta(i,j) = \log \frac{\exp(\mathbf{z}_{f_P,\mathcal{D}(i)} \cdot \mathbf{z}_{f_P,\mathcal{D}(j)} / \tau )}{\sum_{a \in \mathcal{I} \setminus \{i\}} \exp( \mathbf{z}_{f_P,\mathcal{D}(i)} \cdot \mathbf{z}_{f_P,\mathcal{D}(a)}  / \tau )} \nonumber
\end{align}
where $\mathbf{z}_{f_P,x}$ denotes the feature of image $x$ extracted by the \partenc{} $f_P$ after $l_2$-normalization, and $\tau$ is a temperature hyperparameter used to rescale the affinity score.
Minimizing $\superconloss{}(\mathcal{D})$ trains the model to maximize the similarity between features of the same class (positive pair) while pushing away the features from different classes (negative pair). According to \cref{eq:supCon}, as noted in~\cite{khosla2020supervised}, the disagreement between the two features in a positive pair is induced by the variation between image instances and difference resulting from data augmentation.

As an alternative to $\superconloss{}$, the unsupervised contrastive loss $\conloss{}(\mathcal{D})$ shares the same formulation as $\superconloss{}(\mathcal{D})$ while semantic information of class label is excluded. Then, the positive index set for each $i \in \mathcal{I}$ is $\mathcal{I}_{\text{pos}}(i) = \{i+B\}$ for $i \leq B$ and  $\mathcal{I}_{\text{pos}}(i) = \{i-B\}$ for $i>B$, \ie, the disagreement between the two features in a positive pair is only induced by exhaustive augmentation.

In PAL, we use $\superconloss{}$ to train the \partenc{}. $\conloss{}$ is used as an alternative to $\superconloss{}$ for ablation study.
Since $\superconloss{}$  models instance-level similarity between features in positive pairs and push away features of different classes, as shown in our experiments in \cref{section:Discussion}, among the considered alternative variants, the features extracted from $\superconloss{}$-trained \partenc{} facilitates the training of \mainenc{} most.

\subsection{Main Encoder}\label{sec:enc_main}

In this section, we first review the soft-labels introduced in knowledge distillation, and then introduce the alignment constraints at the logit- and feature-level imposed by \partenc{} during \mainenc{} training.

\subsubsection{Knowledge Distillation Preliminary}\label{sec:KD}

As discussed in~\cite{tian2020rethinking}, the teacher model provides soft-labels depicting the fact that some classes are relatively close to each other.
The soft-labels $\mathbf{p} \in \mathcal{R}^{|\mathcal{C}_{base}|}$ are calculated from the output logits $\mathbf{v} \in \mathcal{R}^{|\mathcal{C}_{base}|}$ through softmax operation $\mathbf{p} = \text{Softmax}(\mathbf{v}/\tau)$,
where the temperature $\tau$ can scale the logits and a higher $\tau$ produces a softer probability distribution over base classes.

In addition to the cross-entropy loss on student model, the KL-divergence $\mathcal{L}_{KL}$ is used as objective in teacher-student setting for knowledge distillation and is defined as 
\begin{align}\label{eq:naive_KL}
    & \mathcal{L}_{KL}(\mathbf{p}_{t,x} ||\mathbf{p}_{s,x})  = \sum_{c\in |\mathcal{C}_{base}|} \mathbf{p}_{t,x}(c) \log \frac{\mathbf{p}_{t,x}(c)}{\mathbf{p}_{s,x}(c) } \\
                            &= \sum_{c\in |\mathcal{C}_{base}|} \mathbf{p}_{t,x}(c)\log \mathbf{p}_{t,x}(c) - \sum_{c\in |\mathcal{C}_{base}|} \mathbf{p}_{t,x}(c)\log \mathbf{p}_{s,x}(c) \nonumber \\
                            &= -H(\mathbf{p}_{t,x}) + H(\mathbf{p}_{t,x},\mathbf{p}_{s,x}),\nonumber
\end{align}
where $\mathbf{p}_{t,x}$ and $\mathbf{p}_{s,x}$ are the output probability distribution of the same image $x$ by teacher model and student model. Minimizing $\mathcal{L}_{KL}$ will minimize the cross-entropy $H(\mathbf{p}_{t,x},\mathbf{p}_{s,x})$ between teacher soft-label and student prediction. When the teacher is also trained, the negative entropy of its output $-H(\mathbf{p}_{t,x})$ is minimized. 

Since the teacher model is pretrained to learn hard-anchors for classification, and then predicts logits through the single linear mapping, the logits output is not well-constrained and negative logits with large absolute value exist~\cite{hinton2015distilling}.
As experimentally demonstrated in ~\cite{hinton2015distilling}, a high temperature has to be set for cross-entropy $H(\mathbf{p}_{t,x},\mathbf{p}_{s,x})$ between the student prediction and the soft-label during student model training, 
so that the effect of large negative logits from teacher model can be mitigated and the student model can work better.

Instead of first training a model that has once been rigidly optimized to hard-anchors and then setting a high temperature to reduce impact from large negative logits,
when training a \mainenc{} from scratch with class labels, we propose to use the features extracted by \partenc{} as soft-anchors for providing alignment regularization.
To constrain the logit value of classifier, we first design the classifier as the cosine similarity function between feature representations and class weight vectors.
Then, in addition to minimizing cross-entropy loss for each sample, we use the features of \partenc{} to regularize \mainenc{} and design the constraint method at either logit-level or feature-level alignment.

\begin{figure*}[t]
\begin{center}
   \includegraphics[width=1.0\textwidth]{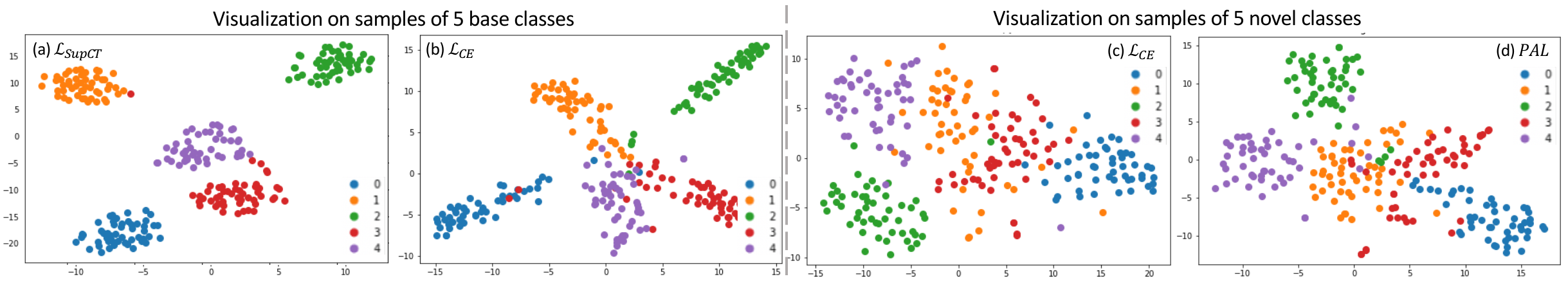}
\end{center}
   \caption{Visualization on five base classes using model trained by (a) supervised contrastive loss $\superconloss{}$ and (b) cross-entropy loss $\mathcal{L}_{CE}$, and visualisation on five novel classes using model learned by (c) $\mathcal{L}_{CE}$ and (d) PAL (our method). (a) $\superconloss{}$ is used to cluster all features using class label, and the features in each cluster can be used as soft-anchors, while (b) $\mathcal{L}_{CE}$ trains network to learn hard-anchors for classification. The feature distribution on novel classes by (d) PAL benefits the prototype classification compared with (c) the distribution of $\mathcal{L}_{CE}$. More visualization can be found in supp. material.}
\label{fig:cluster}
\end{figure*}

\subsubsection{Logit-Level Alignment}

During knowledge distillation, given a query feature of a target class, we calculate affinity score as the dot-product between the query feature and well-trained class weight vector for each base candidate class.
The affinity scores can then be used to describe the relationship between all classes. A candidate class is close to the target class if the corresponding affinity score is high.

Similarly, we use the cosine similarity between the all class weight vectors and the well-trained query features to generate soft labels, and then minimize the cross-entropy,
\[
\mathcal{L}_{logit} = H(\mathbf{p}_{p,x'},\mathbf{p}_{m,x}),
\]
between the soft label $\mathbf{p}_{p,x'}$ generated by \partenc{} and the prediction $\mathbf{p}_{m,x}$ from \mainenc{}, while the image $x'$ and $x$ are of the same class.
As the \partenc{} has been well trained for clustering by maximizing the cosine similarity between features in all positive pairs, we fix the \partenc{} and then extracts features as soft anchors.
Since the \mainenc{} is also trained by maximizing the cosine similarity between the query feature and the class weight vectors, we assume the features by  \textit{Partner} and \mainenc{}s share a common feature space. Thus, we feed features from \partenc{} into the shared classifier and calculate $\mathbf{p}_{p,x'}$.

In terms of implementation,\partenc{} shares the classifier of \mainenc{} to generate cosine similarities as logits. Since the class weight vectors are randomly initialized before training, we adopt a warm-up strategy and increase the weight of $\mathcal{L}_{logit}$ from 0 to 1 as the loss converges and the class weight vectors are gradually learned for each class. 
Notably, our logit-level constraint is different from knowledge distillation and does not minimize the negative entropy of \partenc{}. Since the \partenc{} is not trained to learn hard-anchors, minimizing the negative entropy of \partenc{} is not needed.
Meanwhile, as demonstrated by the ablation study in \cref{tab:ablation_alignment}, since the classifier is randomly initialized at the beginning, comparing with our proposed logit-level alignment ($\text{Row}_{2}$), minimizing the negative entropy of $\mathbf{p}_{p,x'}$ ($\text{Row}_{3}$) will confuse the shared classifier and has negative impact on both classifier and \mainenc{}. 

\begin{table*}[]
\begin{center}
\setlength{\tabcolsep}{0.7em}
{\renewcommand{\arraystretch}{1.0}
\caption{PAL Result on miniImageNet and tieredImageNet datasets. $^\dagger$: Results are generated on Train+Val set}
\begin{tabular}{cccc|cc}
\hlineB{3}
\multirow{2}{*}{Algorithm} & \multirow{2}{*}{Backbone} & \multicolumn{2}{c|}{miniImageNet, 5-way} & \multicolumn{2}{c}{tieredImageNet, 5-way} \\ \cline{3-6} 
                         &                           & 1-shot                  & 5-shot         & 1-shot              & 5-shot              \\ \hline
MAML~\cite{finn2017model}                     & 32-32-32-32               & $48.70 \pm 1.84$           & $63.11 \pm 0.92$  & $51.67 \pm 1.81$       & $70.30 \pm 1.75$       \\
Prototypical Networks~\cite{snell2017prototypical}$^\dagger$   & 64-64-64-64               & $49.42 \pm 0.78$           & $68.20 \pm 0.66$  & $53.31 \pm 0.89$       & $72.69 \pm 0.74$       \\
SNAIL~\cite{SNAIL}                    & ResNet-12                 & $55.71 \pm 0.99$           & $68.88 \pm 0.92$  & -                   & -                   \\
AdaResNet ~\cite{AdaResNet}               & ResNet-12                 & $56.88 \pm 0.62$           & $71.94 \pm 0.57$  & -                   & -                   \\
TADAM~\cite{FC100}                    & ResNet-12                 & $58.50 \pm 0.30$           & $76.70 \pm 0.30$  & -                   & -                   \\
Shot-Free~\cite{ShotFree}                & ResNet-12                 & $59.04 \pm n/a$            & $77.64 \pm n/a$   & $63.52 \pm n/a$        & $82.59 \pm n/a$        \\
TEWAM~\cite{TEWAM}                    & ResNet-12                 & $60.07 \pm n/a$            & $75.90 \pm n/a$   & -                   & -                   \\
MTL~\cite{sun2019mtl}                      & ResNet-12                 & $61.20 \pm 1.80$           & $75.50 \pm 0.80$  & -                   & -                   \\
Variational FSL~\cite{variationalFSL}          & ResNet-12                 & $61.23 \pm 0.26$           & $77.69 \pm 0.17$  & -                   & -                   \\
MetaOptNet~\cite{MetaOptNet}               & ResNet-12                 & $62.64 \pm 0.61$           & $78.63 \pm 0.46$  & $65.99 \pm 0.72$       & $81.56 \pm 0.53$       \\
Fine-tuning~\cite{dhillon2019baseline}              & WRN-28-10                 & $57.73 \pm 0.62$           & $78.17 \pm 0.49$  & $66.58 \pm 0.70$       & $85.55 \pm 0.48$       \\
LEO-trainval~\cite{leo-trainval}$^\dagger$             & WRN-28-10                 & $61.76 \pm 0.08$           & $77.59 \pm 0.12$  & $66.33 \pm 0.05$       & $81.44 \pm 0.09$       \\
Diversity w/Cooperation~\cite{Diversity}            & ResNet-18                 & $59.48\pm0.65$         & $75.62\pm0.48$       & -                   & -                  \\
Associative-Alignment~\cite{a-alignment}            & ResNet-18                & $59.88 \pm 0.67$           & $80.35 \pm 0.73$  & $69.29 \pm 0.56$       & $85.97 \pm 0.49$       \\
AdaMargin~\cite{li2020boosting} & ResNet-12 & $67.10 \pm 0.52$ & $79.54 \pm 0.60$ & - & -\\
DeepEMD~\cite{DeepEMD}            & ResNet-12                & $65.91 \pm 0.82$           & $82.41 \pm 0.56$  & $71.16 \pm 0.87$       & $86.03 \pm 0.58$       \\
MABAS~\cite{MABAS}            & ResNet-12                & $65.08 \pm 0.86$           & $82.70 \pm 0.54$  & -       & -       \\
RFS-simple~\cite{tian2020rethinking}               & ResNet-12                 & $62.02 \pm 0.63$           & $79.64 \pm 0.44$  & $69.74 \pm 0.72$       & $84.41 \pm 0.55$       \\
RFS-distill~\cite{tian2020rethinking}              & ResNet-12                 & $64.82 \pm 0.60$           & $82.14 \pm 0.43$  & $71.52 \pm 0.69$       & $86.03 \pm 0.49$       \\ \hline
PAL (Ours)               & ResNet-12                 & $\mathbf{69.37 \pm 0.64}$  & $\mathbf{84.40 \pm 0.44}$   &  $\mathbf{72.25 \pm 0.72}$       & $\mathbf{86.95 \pm 0.47}$        \\ 
\hlineB{3}
\end{tabular}
\label{tab:imagenet_result}
}
\end{center}
\end{table*}

\subsubsection{Feature-Level Alignment}

Feature-level alignment is achieved by pair-wise comparison between the features extracted by \textit{Partner} and \textit{Main Encoders}.
As stated in \cref{sec:KD}, \mainenc{} is trained with a classifier based on cosine similarity for generating logits and range of logits value is bounded.
Equivalently, learning such a classifier effectively learns a unique anchor for each class and every feature is trained to maximize the cosine similarity with its corresponding class anchor.

During the training process of \partenc{} with $\superconloss{}$, features that belong to the same class are clustered together and separated from features of other classes. Therefore, the clusters of base classes can be considered as a pool and each feature can be considered as a soft-anchor for alignment.
For each image, in addition to the supervised classification signal provided class label, a subset of soft-anchors is sampled from the pools.

Specifically, similar to the idea in supervised contrastive loss, we do pair-wise comparison for feature-level alignment. Given the batch $\mathcal{D}_{M}$ consisting of $|\mathcal{D}_{M}|$ features extracted by \mainenc{}, for each feature instance indexed by $i \in \mathcal{I}_M = \{1...|\mathcal{D}_{M}|\}$, according to class labels, a set of positive features $\mathcal{D}_{P,D_M(i)}^+$ and negative features $\mathcal{D}_{P,D_M(i)}^-$ are randomly sampled from the pools by \partenc{}, such that all features in $\mathcal{D}_{P,D_M(i)}^+$ are of the same class as $\mathcal{D}_{M}(i)$ and every feature in $\mathcal{D}_{P,D_M(i)}^-$ is of a different class from $\mathcal{D}_{M}(i)$. 
Then, we define feature-level constraint as
\begin{align}\label{eq:feat_align}
    &\mathcal{L}_{feat}(\mathcal{D}_{M}) = 
    \sum_{i\in \mathcal{I}_M} \frac{-1}{|\mathcal{D}_{P,D_M(i)}|} \sum_{j\in \{1...|\mathcal{D}_{P,D_M(i)}^+|\}} \Theta(i,j),\\
    &\Theta(i,j) = \log \frac{\exp(D_M(i) \cdot \mathcal{D}_{P,D_M(i)}^+(j) / \tau )}{\sum_{a \in \mathcal{I}_{P,D_M(i)} } \exp(D_M(i) \cdot \mathcal{D}_{P,D_M(i)}(a) / \tau )}, \nonumber
\end{align}
where $\mathcal{D}_{P,D_M(i)} = \mathcal{D}_{P,D_M(i)}^+ \cup \mathcal{D}_{P,D_M(i)}^-$, and $\mathcal{I}_{P,D_M(i)} = \{1... |\mathcal{D}_{P,D_M(i)}|\}$.

\subsubsection{Main Encoder Training}

In summary, in addition to the cross-entropy loss for each sample $\mathcal{L}_{CE} = H(\mathbf{1}_y(x),\mathbf{p}_{m,x})$ where $\mathbf{1}_y(x)$ denotes one-hot vector of class label, we include both logit-level and feature-level alignments in the final training objective of \mainenc{} while the \partenc{} used to extract soft anchors is fixed. In practice, we found that using either $\mathcal{L}_{logit}$ or $\mathcal{L}_{feat}$  provides clear benefit in regularizing the \mainenc{}, whereas summing up both of them produces the best performance. 

\section{Experimental Evaluation}

We evaluated PAL on four benchmark datasets to demonstrate its robustness: miniImagenet~\cite{miniImagenet}, tieredImagenet~\cite{tieredImagenet}, CIFAR-FS~\cite{CIFAR-FS}, and FC100~\cite{FC100}. 
Results are shown in \cref{tab:imagenet_result} and \cref{tab:cifar_result}.
Detailed ablation studies are discussed in \cref{section:Discussion}.

\subsection{Benchmark Datasets \& Training Setup}

\noindent \textbf{Datasets derived from ImageNet~\cite{deng2009imagenet}:} miniImageNet~\cite{miniImagenet,Dynamic_Few_Shot} and tieredImageNet~\cite{tieredImagenet}. 
MiniImageNet~\cite{miniImagenet} contains 100 classes, the class split for (training, few-shot validation, few-shot testing) is (64,16,20). Each base class has 600 images for training and 300 images for fully-supervised classification evaluation~\cite{Dynamic_Few_Shot}. 
TieredImageNet~\cite{tieredImagenet} contains 608 classes with the class split (351,97,160) and around 450K images from base dataset for network training.
All images for the two sets are sized to 84$\times$84.

\noindent \textbf{Dataset derived from CIFAR100:}  CIFAR-FS~\cite{CIFAR-FS} and FC100~\cite{FC100}. 
CIFAR-FS~\cite{CIFAR-FS} contains 100 classes with the class split for (64,16,20).
FC100~\cite{FC100} contains 100 classes with the class split (60,20,20). 
Each class has 600 images and all images for the two sets are of 32$\times$32.

The hierarchical class structure, \ie, some leaf classes can be ground together to a coarse class, are considered for the class split of the TieredImageNet and FC100. 
The leaf classes under the same coarse have more semantic correlation. As there is no overlap of coarse class between the base class set and the novel class set, the adaptation from the base class set to the novel class set will be more challenging.

\noindent \textbf{Training Setup:} On all benchmark datasets, we run experiments using ResNet12~\cite{resnet} as the backbone optimized using stochastic gradient decent (SGD). We use an initial learning rate of 0.03 with a decay factor of 10 starting at the $60$-th epoch, and train for $90$  epochs. The batch size is 64 on Mini-ImageNet, CIFAR-FS and FC100, and is 400 on tiered-ImageNet. The temperature scaling factor $\tau$ in $\superconloss{}$ (\cref{eq:supCon}) and $\mathcal{L}_{feat}$ (\cref{eq:feat_align}) is 0.5 on ImageNet-derived datasets, and is 0.1 on CIFAR-derived datasets. We adopt the data augmentation methods used in SupCT~\cite{khosla2020supervised} and include image rotation prediction for reducing the bias~\cite{gidaris2019boosting}. For each dataset, hyper-parameter settings are the same for both \textit{Partner} and \textit{Main Encoder}.

\begin{table*}[]
\begin{center}
\setlength{\tabcolsep}{0.7em}

{\renewcommand{\arraystretch}{1.0}
\caption{PAL Result on CIFAR-FS and FC100 datasets.}
\begin{tabular}{cccc|cc}
\hlineB{3}
\multirow{2}{*}{Algorithm} & \multirow{2}{*}{Backbone} & \multicolumn{2}{c|}{CIFAR-FS, 5-way}   & \multicolumn{2}{c}{FC100, 5-way} \\ \cline{3-6} 
                          &                           & 1-shot                 & 5-shot        & 1-shot          & 5-shot         \\ \hline
MAML~\cite{finn2017model}                       & 32-32-32-32               & $58.9 \pm 1.9$          & $71.5 \pm 1.0$ & -               & -              \\
Prototypical Networks~\cite{snell2017prototypical}      & 64-64-64-64               & $55.5 \pm 0.7$          & $72.0 \pm 0.6$ & $35.3 \pm 0.6$   & $48.6 \pm 0.6$  \\
Relation Networks~\cite{sung2018learning}          & 64-96-128-256             & $55.0 \pm 1.0$          & $69.3 \pm 0.8$ & -               & -              \\
R2D2~\cite{CIFAR-FS}                       & 96-192-384-512            & $65.3 \pm 0.2$          & $79.4 \pm 0.1$ & -               & -              \\
TADAM~\cite{FC100}                      & ResNet-12                 & -                      & -             & $40.1 \pm 0.4$   & $56.1 \pm 0.4$  \\
Shot-Free~\cite{ShotFree}                  & ResNet-12                 & $69.2 \pm n/a$          & $84.7 \pm n/a$ & -               & -              \\
TEWAM~\cite{TEWAM}                      & ResNet-12                 & $70.4 \pm n/a$          & $81.3 \pm n/a$ & -               & -              \\
Prototypical Networks~\cite{snell2017prototypical}      & ResNet-12                 & $72.2 \pm 0.7$          & $83.5 \pm 0.5$ & $37.5 \pm 0.6$   & $52.5 \pm 0.6$  \\
Boosting~\cite{gidaris2019boosting}                   & WRN-28-10                 & $73.6 \pm 0.3$          & $86.0 \pm 0.2$ & -               & -              \\
MetaOptNet~\cite{MetaOptNet}                 & ResNet-12                 & $72.6 \pm 0.7$          & $84.3 \pm 0.5$ & $41.1 \pm 0.6$   & $55.5 \pm 0.6$  \\
Associative-Alignment~\cite{a-alignment}                 & ResNet-18                 & -          & - & $45.8 \pm 0.5$   & $59.7 \pm 0.6$  \\
DeepEMD~\cite{DeepEMD}                 & ResNet-12                 & -          & - & $46.5 \pm 0.8$   & $63.2 \pm 0.7$  \\
MABAS~\cite{MABAS}                 & ResNet-12                 & $73.5 \pm 0.9$   & $85.5 \pm 0.7$ & $42.3 \pm 0.8$   & $57.6 \pm 0.8$  \\
RFS-simple~\cite{tian2020rethinking}                 & ResNet-12                 & $71.5 \pm 0.8$          & $86.0 \pm 0.5$ & $42.6 \pm 0.7$   & $59.1 \pm 0.6$  \\
RFS-distill~\cite{tian2020rethinking}                & ResNet-12                 & $73.9 \pm 0.8$          & $86.9 \pm 0.5$ & $44.6 \pm 0.7$   & $60.9 \pm 0.6$  \\ \hline
PAL (Ours)                 & ResNet-12                 & $\mathbf{77.1 \pm 0.7}$ & $\mathbf{88.0 \pm 0.5}$  & $\mathbf{47.2 \pm 0.6}$    & $\mathbf{64.0 \pm 0.6}$   \\
\hlineB{3}
\end{tabular}
\label{tab:cifar_result}
}
\end{center}
\end{table*}

\subsection{Comparison With State-Of-The-Art}

We compare the performance of PAL with state-of-the art (SOTA) methods.
The idea of multi-task training is adopted in \cite{FC100} and combines the objectives of classification task and 5-way few-shot task during training.
Similarly, a strong baseline is first obtained through pretraining, and the idea of transfer learning is then used to perform training on hard-task~\cite{sun2019mtl} or through finetuning~\cite{dhillon2019baseline}.
Recently, knowledge distillation has been implemented by~\cite{tian2020rethinking} and improves the performance on few-shot tasks clearly. 
Despite the success of previous work, PAL outperforms SOTA methods on all four benchmark datasets in both 1-shot and 5-shot scenarios, which demonstrates the advantages of the proposed PAL learning scheme, where we train \partenc{} and \mainenc{} under different objective and impose constraint by logit-level alignment and feature-level alignment.
Besides, AdaMargin~\cite{li2020boosting} introduces external structured semantic knowledge to model the relationship between classes, and meta-learn a discriminative feature space. However, our method still achieves better result on the few-shot tasks.
Furthermore, by comparing the performance boost between mini-ImageNet and tieredImagenet, and between CIFAR-FS and FC100, we notice that PAL's advantage could be better revealed on the dataset without hierarchical structure (mini-ImageNet, CIFAR-FS), but can still facilitate the few-shot tasks with hierarchical structure (tieredImagenet and FC100).

\subsection{Discussion}
\label{section:Discussion}

Partner-Assisted Learning involves supervised contrastive loss and cross-entropy loss during training. A clear uni-direction from the \partenc{} \partfunc{} to the \mainenc{} \mainfunc{} is set by sampling features of \partfunc{} as soft anchors to assist the training of \mainfunc{} in addition to class labels.
To this end, we study the alternatives of PAL by (1) altering the integration direction of the two objective types and (2) changing the objective of \partenc{} training. We also studied (3) the impact of different alignment losses, which serves as a comparison with knowledge distillation.

\noindent \textbf{Integration of Objectives:} 
PAL uses $\superconloss{}$ to pretrain \partfunc{} to extract soft-anchors, which are then used to regularize \mainfunc{} trained by $\mathcal{L}_{CE}$, \ie, $\superconloss{} \rightarrow \mathcal{L}_{CE}$. We study the variants by altering the integration direction and have
\begin{itemize}[noitemsep,topsep=0pt, leftmargin=*]
\item  Uni-direction  $\mathcal{L}_{CE} \rightarrow \superconloss{}$,  $\superconloss{} \rightarrow \mathcal{L}_{CE}$,
\item  Mutual-learning $\superconloss{}\leftrightarrow\mathcal{L}_{CE}$: contemporarily train two networks from scratch and use the model under $\mathcal{L}_{CE}$ for evaluation,
\item  Multi-task learning on one network: $\superconloss{}$+$\mathcal{L}_{CE}$,
\item  Single-Objective training: $\mathcal{L}_{CE}$, $\superconloss{}$.
\end{itemize}

As shown in \cref{tab:objective_integ}, we observe that the network trained using PAL ($\superconloss{} \rightarrow \mathcal{L}_{CE}$) generalizes best with a clear margin on the few-shot classification on novel classes.
On the base classes test samples, PAL continues to achieve high top-1 accuracy that is very close to the best score by multi-task learning, model trained with  ($\superconloss{}$+$\mathcal{L}_{CE}$), and PAL clearly outperforms the rest methods. 
Meanwhile, compared to the two single-objective methods, even though the model trained by $\superconloss{}$ is not as strong as the model trained by $\mathcal{L}_{CE}$, it can still be used to regularize the training under $\mathcal{L}_{CE}$ and boost the performance.

\begin{table}[]
	\renewcommand{\arraystretch}{0.9}
	\caption{Ablation study of mini-ImageNet on the performance by different training scheme combining two objectives. Our method works best on the few-shot classification.}
	\resizebox{\linewidth}{!}{
		\begin{tabular}{c cc c}
			\hlineB{3}
			\multirow{2}{*}{\shortstack{Train Scheme}} & \multicolumn{2}{c}{5-Way Few-shot} & \multirow{2}{*}{Base} \\ \cline{2-3}
			& 1-Shot & 5-Shots & \\ \hline
			$\mathcal{L}_{CE}$												   & $63.76$          & $81.17$            & $80.90$                         \\
			$\mathcal{L}_{SupCT}$											& $62.29$           & $76.32$            & $n/a$                         \\
			$\mathcal{L}_{CE}$ + $\superconloss{}$					  & $67.53$          & $82.14$            & $\mathit{83.20}$                         \\
			$\superconloss{} \leftrightarrow \mathcal{L}_{CE} $	& $65.21$           & $81.53$            & $80.13$          \\
			$\mathcal{L}_{CE} \rightarrow \superconloss{}$		 & $66.54$           & $81.83$            & $80.39$                         \\
			$\mathbf{\superconloss{} \rightarrow \mathcal{L}_{CE}}$ [ours]		& $\mathbf{69.37}$           & $\mathbf{84.40}$            & $82.98$                         \\ 
			\hlineB{3}
			\multicolumn{4}{l}{$\ast ~ \text{Single objective}~\mathcal{L}_{SupCT}$ don't train a base classifier.}
		\end{tabular}
		\label{tab:objective_integ}
	}
\end{table}

\noindent \textbf{Impact of \partenc{} on \mainenc{}} is studied by evaluating the performance of \mainenc{}, \ie, accuracy of few-shot tasks and conventional classification task on base test data. We select \partenc{}s trained by $\mathcal{L}_{CE}$, $\mathcal{L}_{CT}$, and $\superconloss{}$ for comparison. Both $\mathcal{L}_{\text{feat}}$ and $\mathcal{L}_{\text{logit}}$ are used for all three methods for fair comparison.

As shown in \cref{tab:main_by_dif_partner}, the alignment losses consistently introduce performance improvements on few-shot tasks, and our method ($\superconloss{} \rightarrow \mathcal{L}_{CE}$) continues to achieve the best performance on both few-shot and fully-supervised tasks. 
Since $\mathcal{L}_{CT}$ does not utilize the class label information during training, the extracted soft-anchors are not as discriminative as the other two methods and the performance improvement is limited. 
Meanwhile, if both the \mainenc{} and the \partenc{} are trained under the $\mathcal{L}_{CE}$, the features are mainly learned w.r.t hard-anchors. Therefore, the knowledge is similar for the two networks and the performance improvement is limited.
In contrast, supervised contrastive loss includes both the label information and pair-wise similarity comparison among all features, so that it could preserve more details to facilitate the training of \mainenc{}.

\noindent \textbf{Alignment versus Knowledge Distillation:} 
Although teacher-student model with knowledge distillation can train a good feature extractor for few-shot prototype classification, a good teacher is a critical prerequisite for training the student. As discussed in \cref{sec:KD}, the negative entropy of teacher model is minimized if the teacher model is also tuned during distillation.
To this end, as shown in \cref{tab:ablation_alignment}, we quantitatively investigate the effectiveness of both $\mathcal{L}_{\text{feat}}$ and $\mathcal{L}_{\text{logit}}$ by applying various combinations. We also compare with the performance under teacher-student setting and set the KL-divergence as the logit-level loss. 

By comparing $\text{Row}_{1,2,4,5}$, using either $\mathcal{L}_{\text{feat}}$ or $\mathcal{L}_{\text{logit}}$ can regularize the training of \mainfunc{}, and clearly improve performance on both few-shot and fully-supervised tasks. Using $\mathcal{L}_{\text{feat}}$ only can even outperform the multi-task training ($\text{Row}_3$ in \cref{tab:objective_integ}). Meanwhile, adding these two losses can achieve the best scores on few-shot tasks.
By comparing $\text{Row}_{2,3}$ or $\text{Row}_{5,6}$, minimizing the negative entropy of outputs from \partfunc{} has negative impact on the training of \mainfunc{}. As the \partfunc{} is pretrained to perform pair-wise comparison, the distance between all features and all feature clusters has already been modeled. Since \partfunc{} is not trained to do classification, the details irrelevant to classification on base domain could be preserved. 
Therefore, minimizing the negative entropy of \partfunc{} may make probability output of \partfunc{} less sensitive to the instance-level distances and then embed the soft-anchors with more uncertainty. 
Even though the \partenc{} (trained under $\superconloss{}$) is not as strong as the teacher model (trained under $\mathcal{L}_{CE}$) in few-shot tasks, it can provide meaningful soft-anchors to assist the training of \mainenc{} in our proposed PAL framework.

\noindent \textbf{PAL versus Mutual Learning:} Mutual learning trains two peer networks jointly from scratch either under the same task~\cite{zhang2018deep} or different tasks ($\text{Row}_4$ in \cref{tab:objective_integ}). The alignment between probability distribution outputs is applied~\cite{zhang2018deep}. Even though the two networks are expected to learn in different direction, the optimization direction is still not clearly modeled in such framework. In contrast, our method sets the features extracted by \partenc{} as soft-anchors to model the distance of positive pair and negative pair, and then train \mainenc{} through alignment.

\begin{table}[]
\begin{center}
\caption{Ablation study of mini-ImageNet on the \mainenc{} performance affected by different \partenc{}.The \partenc{} trained by $\superconloss{}$ benefits the training of \mainenc{} most.}
\resizebox{\linewidth}{!}
{
    \begin{tabular}{l c c c}
    \hlineB{3}
    \multirow{2}{*}{Loss of \partfunc{}} & \multicolumn{2}{c}{5-Way Few Shot} & \multirow{2}{*}{Base} \\ \cline{2-3}
     &  1-Shot  &   5-Shots           &   \\ \hline
    - & $63.76 \pm 0.62$ & $81.17 \pm 0.45$ & $80.90$ \\
    $\mathcal{L}_{CT} \rightarrow \mathcal{L}_{CE}$ & $65.89 \pm 0.67$ & $80.84 \pm 0.48$ & $80.66$ \\
    $\mathcal{L}_{CE} \rightarrow \mathcal{L}_{CE}$ & $66.95 \pm 0.65$ & $81.54 \pm 0.48$ & $81.45$ \\
    $\superconloss{}  \rightarrow \mathcal{L}_{CE}$ & $69.37 \pm 0.64$  & $84.40 \pm 0.44$ & $82.98$ \\
    \hlineB{3}
    \end{tabular}
}
\label{tab:main_by_dif_partner}
\end{center}
\end{table}

\begin{table}[]
\setlength{\tabcolsep}{0.5em}
\caption{Ablation study on mini-ImageNet on alignment losses. Both feature-level and logit-level losses facilitates the training of \mainenc{} consistently.}
\resizebox{\linewidth}{!}{
    \begin{tabular}{c c | c c | c}
    \hlineB{3}
    \multicolumn{2}{c|}{Alignment losses} & \multicolumn{2}{c|}{5-way few-shot} & \multirow{2}{*}{Base}\\ \cline{1-4}
    feature & logit & 1-shot &  5-shots &  \\ \cline{1-5}
     - & - & $63.76 \pm 0.62$ & $81.17 \pm 0.45$ & $80.90$\\
     - & $\mathcal{L}_{logit}$ & $66.35 \pm 0.63$ & $81.32 \pm 0.46$ & $82.50$ \\
     - & $\mathcal{L}_{KL}$ & $64.76 \pm 0.62$ & $80.58 \pm 0.47$ & $81.98$ \\
     $\mathcal{L}_{feat}$ & - & $68.03 \pm 0.63$ & $83.38 \pm 0.44$ & $\mathbf{83.94}$\\
     $\mathcal{L}_{feat}$ & $\mathcal{L}_{logit}$ & $\mathbf{69.37 \pm 0.64}$  & $\mathbf{84.40 \pm 0.44}$ & $\mathit{82.98}$\\
     $\mathcal{L}_{feat}$ & $\mathcal{L}_{KL}$ & $68.75 \pm 0.62$ & $82.83 \pm 0.45$ & $83.40$\\
    \hlineB{3}
    \end{tabular}
}
\label{tab:ablation_alignment}
\end{table}

\section{Conclusion}
In this paper, we have proposed the \textit{Partner Assisted Learning} (PAL) to obtain an essential feature extractor for few-shot classification. 
We pre-train the \partenc{} with supervised contrastive learning to obtain soft-anchors. 
Then, we fix the \textit{partner} model and impose the constraints at either feature-level or logit-level to train the \mainenc{} from scratch while seeking for classification.
With the \textit{main} model, both the classification accuracy on novel classes (few-shot) and on base classes (large-scale) are improved.
Detailed ablation study is performed to compare potential variants of PAL and our method outperforms clearly all variants on few-shot tasks.
Experiments on four benchmarks demonstrate the effectiveness of our method. 

\section*{Acknowledgements}
{\small
\begin{spacing}{0.9}
\noindent This material is based on research sponsored by Air Force Research Laboratory (AFRL) under agreement number FA8750-19-1-1000. 
The U.S. Government is authorized to reproduce and distribute reprints for Government purposes notwithstanding any copyright notation therein. 
The views and conclusions contained herein are those of the authors and should not be interpreted as necessarily representing the official policies or endorsements, either expressed or implied, of Air Force Laboratory, DARPA or the U.S. Government.
\end{spacing}
}

{\small
\bibliographystyle{ieee_fullname}
\bibliography{reference}
}

\end{document}